\pdfoutput=1

\documentclass[a4paper,11pt]{article}

\usepackage[preprint]{acl}

\usepackage{times}
\usepackage{latexsym}

\usepackage[T1]{fontenc}
\usepackage[utf8]{inputenc}
\usepackage{multirow}

\usepackage{microtype} 

\usepackage{inconsolata} 

\usepackage{subfiles}

\usepackage[most]{tcolorbox}
\usepackage{booktabs} 
\usepackage{enumitem}

\usepackage{lmodern}

\usepackage{amsmath}
\usepackage{amsthm}
\usepackage{amssymb}

\usepackage[capitalise]{cleveref}

\usepackage{graphicx}

\usepackage{xcolor}
\usepackage{hyperref}

\usepackage{caption}

\usepackage{siunitx} 
\usepackage{comment}

\usepackage{csquotes} 
\usepackage{yfonts,color} 
\usepackage{lettrine}
\usepackage{GoudyIn}

\LettrineTextFont{\itshape}
\theoremstyle{definition}

\definecolor{myblue}{RGB}{20,80,150}  
\definecolor{myred}{RGB}{160,30,30}   
\definecolor{mygreen}{RGB}{50,120,50} 

\title{Once Upon a Time:\\Interactive Learning for Storytelling with Small Language Models}

\author{
    Jonas Mayer Martins ~~~~
    Ali Hamza Bashir \\
    \textbf{~ Muhammad Rehan Khalid ~~~~~
    Lisa Beinborn} \vspace{2mm} \\ 
    University of Göttingen, Institute of Computer Science, Germany \\ 
    \texttt{firstname.lastname@uni-goettingen.de} \\}

\begin{document}

\maketitle

\begin{abstract}
    Children efficiently acquire language not just by listening, but by interacting with others in their social environment. Conversely, large language models are typically trained with next-word prediction on massive amounts of text. Motivated by this contrast, we investigate whether language models can be trained with less data by learning not only from next-word prediction but also from high-level, cognitively inspired feedback. We train a student model to generate stories, which a teacher model rates on readability, narrative coherence, and creativity. By varying the amount of pretraining before the feedback loop, we assess the impact of this interactive learning on formal and functional linguistic competence. We find that the high-level feedback is highly data efficient: With just 1\,M words of input in interactive learning, storytelling skills can improve as much as with 410\,M words of next-word prediction.
\end{abstract}

\begin{center}
\small
\href{https://huggingface.co/llm-slice}{%
  \raisebox{-0.4em}{\includegraphics[height=1.3em]{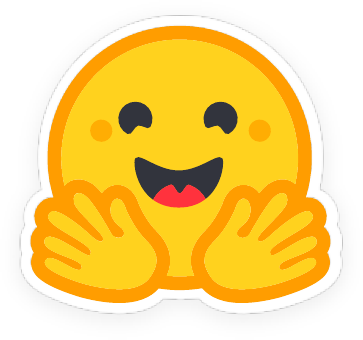}}
  Models and data} 
\hspace{8pt}|\hspace{7pt}
\href{https://gitlab.gwdg.de/huds/projects/blm2025/-/tree/e7dc9b43c902e76a5ed75335c70fc8e768d31aee/}{%
  \raisebox{-0.33em}{\includegraphics[height=1.2em]{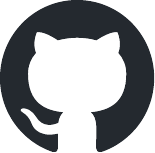}} 
  Code repository}
\end{center}

\section{Introduction}


\lettrine{H}{umans} are storytelling animals \cite{Gottschall2012,Campbell2008}. From early myths to modern science, narratives have served not only as entertainment but also as cognitive tools to make sense of the world. Scientific models and historical accounts, personal and collective identities, and even abstract institutions such as currency, law, and national borders can all be understood as shared stories \cite{Bruner1991}. Through our capacity for language, we establish a communicative common ground to align intentions, construct shared realities, and thus cooperate at societal scales \cite{Tomasello2008,Tomasello2014,ClarkS1989,ClarkB1991}.

In recent years, language models have achieved surprising proficiency in generating natural language. However, training these artificial neural networks with billions to trillions of parameters is inefficient \cite{WilcoxHMLW2024}. While modern supercomputers are trained on the order of $10^{13}$ words \cite{DeepSeekAI2025}, a child is exposed to between $10^8$ and $10^9$ words by age 13, extrapolating from \citet{GilkersonRWMG2017}. How do children acquire language so efficiently? In this work, we explore one potential ingredient: enriching the learning signal for language models beyond classical next-word prediction \cite{StoplerANCW2025}. 

Artificial and biological neural networks differ in structure and dynamics, yet both can acquire complex linguistic behavior \cite{EvansonLK2023}. The standard training objective for language models---next-word prediction---superficially resembles predictive processing \cite{Clark2013,RyskinN2023}, but does not reflect the rich, interactive learning experienced by children. We hypothesize that incorporating high-level feedback can guide language models toward more efficient functional linguistic competence, i.e., coherent, pragmatic, and creative use of language \cite{MahowaldIBKT2024}.

\begin{figure*}[htp]
    \centering
    \includegraphics{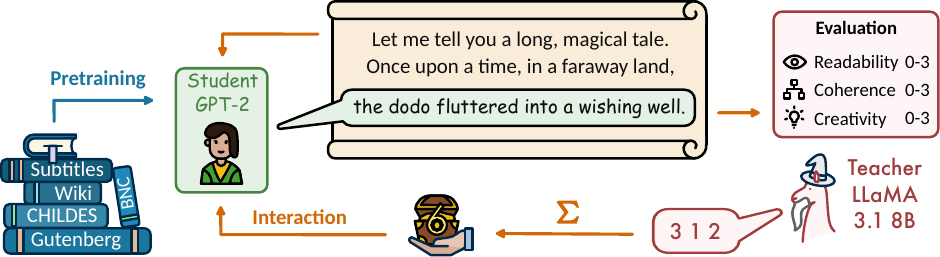}
    \caption{Schematic of the interactive learning setup with storytelling feedback. During pretraining, the student model optimizes next-word prediction on the BabyLM corpus. In the interaction stage, the student model completes a story prompt. A teacher model then evaluates the story on three criteria using a Likert scale from 0 to 3. The student receives the sum of these scores as a reward and updates its parameters to generate stories that maximize the expected reward.}
    \label{fig:graphical_abstract}
\end{figure*}
While the human brain excels at finding patterns in sensory input---a capacity central to early language learning \cite{Saffran2020}---children are more than just passive recipients of this input. Instead, they learn language in a social context, shaped by interaction and feedback from caregivers \cite{Tomasello2008,Clark2018}. This feedback includes both implicit cues, such as contingent responses and repetitions, and explicit forms, such as corrections and confirmations \cite{CheathamJP2015,NikolausF2023}.

By contrast, traditional language modeling is fully self-supervised. External feedback is integrated only later, during fine-tuning for applied tasks, when the model receives feedback from labeled examples \cite{ParthasarathyZKS2024}. More recently, reinforcement learning (RL) has been introduced to language modeling to better align model outputs with human preferences.

In this work, we replace part of the next-word prediction in pretraining by reinforcement learning in interaction with a teacher model, employing storytelling as a task that requires functional linguistic competence, see \cref{fig:graphical_abstract}. After pretraining on the BabyLM corpus \cite{CharpentierCCGH2025}, the student model enters the interaction loop: First, the student generates a story from a generic snippet. Next, the teacher model judges the generated story with respect to readability, narrative coherence, and creativity. Finally, the student model receives the sum of the teacher scores as a reward and updates its parameters to maximize the expected reward.

We assess how high-level narrative and linguistic feedback impacts the student model's learning dynamics. Specifically, we demonstrate that partially replacing next-word prediction with interaction augments storytelling ability without compromising low-level linguistic generalization. Remarkably, with less than 1\,M input words of interactive learning, storytelling skills improve as much as 410\,M additional words of conventional pretraining. Finally, we examine how the amount of pretraining influences the effectiveness and dynamics of reinforcement learning for storytelling.

\section{Interactive learning for small language models}
\label{sec:related_work}

Prior work on data efficiency in language modeling motivates alternative training objectives. Discussing storytelling as a lens for evaluating linguistic competence, we present interactive learning as a cognitively inspired approach to improving data efficiency and functional language skills in small models.

\subsection{Scaling and parsimony}

Large language models generally perform better with more parameters and more training data \cite{BahriDKLS2024}. From a cognitive perspective, data parsimony is of particular interest. A child encounters orders of magnitude fewer words than large language models: Extrapolating from \citet{GilkersonRWMG2017}, we estimate that by age 13 a child has been exposed to around 100 million to 1~billion words---only a fraction of the input given to modern language models.
Inspired by how children acquire language,\footnote{We use language \emph{learning} and \emph{acquisition} interchangeably in this work.} the BabyLM Challenge seeks to close this gap in data efficiency \cite{WarstadtMCWZ2023,HuMRWL2024,CharpentierCCGH2025}.
The findings from previous BabyLM challenges show that the most promising improvements in model performance come from changes in architecture and training objective \cite{WarstadtMCWZ2023,HuMRWL2024}. 

We hypothesize that the next-word prediction objective---operating at the word or subword level---is too fine-grained to foster sufficient abstraction. In addition, next-word prediction requires multiple exposures to each word for effective learning and introduces frequency biases and anisotropy in model representations \cite{diehl-martinez-etal-2024-mitigating,godey-etal-2024-anisotropy}. Achieving greater data efficiency may require a more comprehensive signal that incorporates high-level feedback.

\subsection{Modeling storytelling}

Humans communicate through stories and improve as storytellers by learning from interactive feedback. As a learning objective, storytelling is particularly valuable because it requires \emph{functional linguistic competence} (i.e., pragmatic use of language in real-world situations), as opposed to \emph{formal linguistic competence} (i.e., knowledge of linguistic rules and patterns) \cite{MahowaldIBKT2024}. However, what defines a good story is difficult to formalize \cite{ChhunCSC2022} and existing metrics align poorly with human judgments \cite{GuanZFLD2021}. 

Contemporary language models can produce fluent and grammatically correct stories but frequently struggle with coherence, creativity, and narrative structure \cite{SeePSYM2019,xie-etal-2023-next}. For example, models often fail at \emph{entity tracking} (keeping track of facts about the world in a story), which is crucial for coherent stories \cite{KimS2023,LiNA2021}. We propose that these functional skills can be improved by enriching the training objective with storytelling feedback.

\subsection{Interactive learning}

In multi-agent signaling games, the interactions of agents can lead to the emergence of novel communication protocols or even languages \cite{BoldtM2024,BernardMT2024,lazaridou-etal-2020-multi}. Also, cognitively inspired feedback can improve model performance \cite{NikolausF2021,SahaHB2023,StoplerANCW2025}. While previous work has explored various forms of feedback, our approach lets the student model generate stories freely in response to a writing prompt, while the teacher model provides high-level feedback on story quality.

Reinforcement learning, although a well-established method in machine learning, is relatively new to natural language processing \cite{ParthasarathyZKS2024,HavrillaDRND2024}. With regard to storytelling, reinforcement learning of sufficiently pretrained models appears surprisingly robust to sparse reward signals \cite{zhao-etal-2023-babystories,WuWZH2025}. Unlike knowledge distillation, which approximates the function of a large language model through a model with fewer parameters \cite{DasguptaCB2023}, our method uses textual feedback rather than probability distributions. This approach may be less computationally efficient, but it provides a more developmentally plausible reward signal, emulating student-teacher or child-caregiver interaction.

\section{Methodology}

As illustrated in \cref{fig:graphical_abstract}, we model interaction as follows: A pretrained \emph{student model} generates a story, which a \emph{teacher model} then rates based on \emph{evaluation instructions}. The teacher's scores serve as the reward signal for reinforcement learning via proximal policy optimization (PPO) \cite{ParthasarathyZKS2024}.

\paragraph{Baselines}
We compare the student model against two baselines from the 2025 BabyLM challenge \cite{CharpentierCCGH2025}:

\textbf{\href{https://huggingface.co/BabyLM-community/babylm-baseline-100m-gpt2}{\texttt{1000M-pre}} baseline}: trained on $100\,\mathrm{M}$ unique words of the BabyLM corpus for 10~epochs with next-word prediction.

\textbf{\href{https://huggingface.co/BabyLM-community/babylm-interaction-baseline-simpo}{\texttt{SimPO}} baseline}: trained for 7~epochs with next-word prediction on the BabyLM corpus and 2~epochs interleaving next-word prediction with reinforcement learning. The reward is based on how similar the story completions of the student are to that of the teacher, providing corrective feedback.

\paragraph{Student model} For our experiments, we use the same \texttt{GPT-2-small} architecture as the baseline for the student model and similar hyperparameters, see \cref{app:model_parameters:pretraining}. We divide the training into two stages: 

\textbf{\texttt{900M-pre} baseline}: To stay within a word budget of 100\,M words per epoch, we pretrain first on $90\,\%$ of the 100\,M BabyLM corpus for 10~epochs.

\textbf{\texttt{900M-RL} model}: Subsequently, we do interactive learning with 1\,M words of input. This yields fewer input words to the student model than the other baselines, namely, 901\,M and 1,000\,M words, respectively.

\paragraph{Teacher model}

Evaluating the quality of a story is a difficult task that requires both accurate judgments and computational efficiency. Based on pilot experiments, we select \texttt{Llama~3.1~8B Instruct} \cite{GrattafioriDJPK2024}.\footnote{Out of the three Llama Instruct models available for the Interaction Track of the BabyLM challenge (3.1~8B, 3.2~3B, and 3.2~1B), the largest one (Llama 3.1~8B Instruct) provides story scores with a reasonably high signal-to-noise ratio that aligned best with the developers' assessments of the story.}

To mirror the student-teacher analogy, we keep the teacher model fixed throughout training.

\paragraph{Story generation}

To obtain a viable reward signal in reinforcement learning, we must elicit story-like outputs from the student model. We use the archetypal storytelling opening:
\begin{tcolorbox}[colback=mygreen!10, colframe=mygreen!80, title={Student Model Input}]
\emph{Let me tell you a long, magical tale.\\
Once upon a time, in a faraway land,}
\end{tcolorbox}

\paragraph{Teacher feedback}

Defining the quality of a story is notoriously challenging \cite{ChhunCSC2022}.
Following \citet{GuanZFLD2021}, we let the teacher model evaluate the student story on three criteria: readability, narrative coherence, and creativity.

Careful optimization of the teacher instructions was required for a strong and accurate learning signal, as language models are often highly sensitive to prompt phrasing \cite{ChhunCSC2022} and prone to label-induced biases \cite{SarafBBAB2025}. During development, we refined the instructions to discourage shortcutting and ensure alignment with human judgment. We use rubrics to anchor the teacher's responses and provide examples of expected outputs. For each criterion, the teacher assigns a score from $0$ (worst) to $3$ (best), yielding robust and concise feedback. The full evaluation instructions are given in \cref{app:teacher_prompt}.

\paragraph{Reward}

We use PPO to optimize the language model's policy for maximum expected reward.
The reward $R$ is calculated by combining the teacher scores $s_i \in \{0,1,2,3\} $ for the three criteria $i$ with a story length incentive based on the number of generated words $L$: 
\begin{equation}
    R = \frac{1}{1 + \alpha}
        \bigg[
        \frac{1}{9} \sum_{i=1}^3 s_i
      + \alpha \,\frac{L}{L_{\max}}
        \bigg]
      + r_\mathrm{KL}\,,
\end{equation}
$L_\mathrm{max}=100$ is the maximum allowed number of subword tokens (to normalize length), and $\alpha=0.4$ controls the relative weight of the length bonus. The Kullback--Leibler (KL) divergence $r_\mathrm{KL}$ prevents the trained model from diverging too far from the pretrained baseline. See \cref{app:model_parameters:RL} for full training parameters.

\paragraph{Experimental setup}

We first pretrain the \texttt{GPT-2-small} student model on 90\,\% of the BabyLM corpus for 10 epochs. To track the learning dynamics, we save checkpoints at logarithmically spaced intervals (1\,M, 2\,M, \dots, 10\,M, 20\,M, \dots, 100\,M, 200\,M, \dots, and 900\,M words seen by the model). The final checkpoint constitutes our \texttt{900M-pre} baseline.

To assess the amount of pretraining necessary for efficient RL, we start the reinforcement learning from selected checkpoints (20\,M, 50\,M, 90\,M, 200\,M, 500\,M, 900\,M)\footnote{The tags (e.g., 900\,M) refer to the number of pretrained words, not model size.} and train for 1\,M words in 331.2k interactions (that is, 331.2k stories told), with evaluation checkpoints every 100k words.

During reinforcement learning, we log the stories, story length, teacher scores, as well as the KL divergence. The figures in \cref{sec:results:storytelling} report Gaussian-smoothed batch averages ($\sigma = 30$ with batch size 360), unless otherwise noted.

\begin{figure*}[htb]
    \centering
    \includegraphics[width=\linewidth]{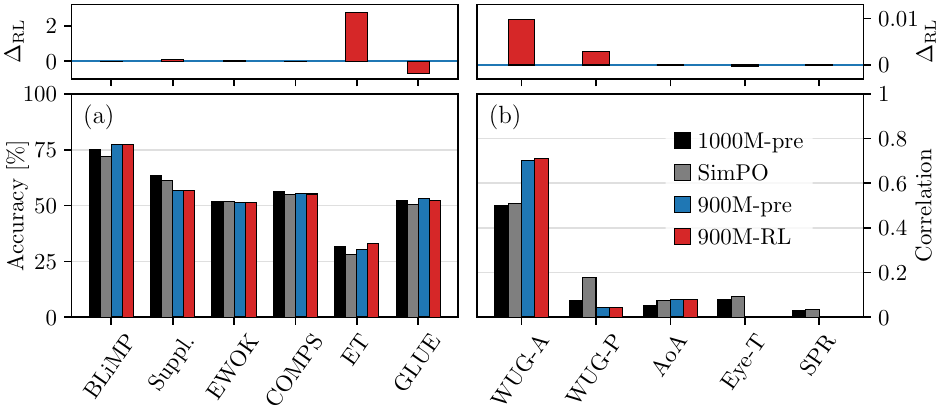}
    \caption{Evaluation on the BabyLM tasks, cf.\ \cref{tab:evaluation_setup}. The bottom panels show the (a) accuracy and (b) correlation or partial correlation on the respective tasks for the next-word prediction baseline \texttt{1000M-pre}, the interaction baseline \texttt{SimPO}, and the model before and after interactive reinforcement learning (RL). The top panels indicate the difference of the 900\,M model before and after interaction learning (in percentage points on the left, correlation differences on the right). \texttt{GLUE} encompasses all fine-tuning tasks.}
    \label{fig:blm_eval}
\end{figure*}

\section{Evaluation setup}
We use the \href{https://github.com/babylm/evaluation-pipeline-2025}{evaluation pipeline} of the 2025 BabyLM Challenge \cite{CharpentierCCGH2025}. It comprises nine zero-shot diagnostic benchmarks and seven task-specific datasets that require model fine-tuning (see \cref{app:BLM_evals}). 

\paragraph{Zero-shot diagnostics} This suite evaluates the linguistic and conceptual capabilities of the language model by comparing its language modeling probabilities to human judgments. Minimal pair tasks are used to assess whether the model assigns higher probability to the more acceptable sentence. Each pair consists of two minimally contrastive sentences that isolate a certain phenomenon relating to syntactic and semantic grammaticality (\texttt{BLiMP}), dialogue and question processing (\texttt{BLimP supplement}), world knowledge about physical and social concepts (\texttt{EWoK}), and property inheritance (\texttt{COMPS}). In addition, the probabilities are correlated with human ratings for morphological properties of pseudo-words (\texttt{WUGs}), and to age-of-acquisition labels (\texttt{AoA}). Context integration capabilities of the model are tested by evaluating the proportion of the variance in eye-tracking (\texttt{Eye-T}) and self-paced reading (\texttt{SPR}) signals that is predictable from the surprisal of the model and by the accuracy of predicting the final state of an entity (entity tracking, \texttt{ET}) after a series of operations described as natural language discourse.

\paragraph{Task-specific fine-tuning} The applicability of the model for downstream tasks is evaluated by its task-specific accuracy after supervised fine-tuning for question answering (\texttt{BoolQ} and \texttt{MultiRC}), natural language inference (\texttt{MNLI} and \texttt{RTE}), paraphrase recognition (\texttt{MRPC} and \texttt{QQP}), and coreference resolution (\texttt{WSC}). In the results, the fine-tuning tasks are summarized as \texttt{GLUE}. See \cref{app:model_parameters:fine-tuning} for fine-tuning parameters.

\section{Results and discussion}
We first examine the effect of our interaction model on formal linguistic competence as assessed by the BabyLM evaluation pipeline. We then analyze how storytelling skills improve through reinforcement learning, and explore the training dynamics.

\subsection{Formal linguistic competence}
\label{sec:results:blm_eval}

We evaluate formal linguistic competence using the BabyLM tasks, comparing our model pretrained on 900\,M words before (\texttt{900M-pre}) and after (\texttt{900M-RL}) interactive reinforcement learning. We also compare with a baseline pretrained on 1,000\,M words (\texttt{1000M-pre}), and an interaction baseline with a different training objective (\texttt{SimPO}). The results are summarized in \cref{fig:blm_eval}; for detailed values, see \cref{tab:BLM_eval_results}.

We observe that the two baselines, \texttt{1000M-pre} and \texttt{900M-pre}, achieve similar performance on most tasks. This suggests that the missing $10\%$ of the pretraining corpus and thus 100\,M additional words in pretraining have little effect on formal linguistic competence.

Strikingly, as shown in the top panels in \cref{fig:blm_eval}, the accuracy on entity tracking (\texttt{ET}) increases the most, from $30.3\,\%$ to $33.1\,\%$, and correlations on the two \texttt{WUG} tasks improve marginally. Although the teacher reward was not tailored to any of these tasks, improved entity tracking likely reflects the importance of maintaining narrative coherence---specifically, keeping track of characters and objects---in storytelling. Accuracy on the \texttt{GLUE} benchmark drops slightly by $0.7$ percentage points after interaction. Notably, interactive reinforcement learning does not affect most other BabyLM tasks. 

The \texttt{SimPO} baseline, despite being exposed to more words during interaction, does not differ much from the baselines and performs slightly worse than \texttt{1000M-pre} on \texttt{BLiMP}, \texttt{BLiMP} Supplement, \texttt{ET}, and \texttt{GLUE}. 

As shown in panel~(b) the metrics measuring alignment with psycholinguistic data (\texttt{AoA}, \texttt{Eye-T}, and \texttt{SPR}) have less consistent trends than the accuracy-based scores in panel~(a) and the correlations of all models are below 0.1. The \texttt{WUG-A} task has a high correlation between 0.5 and 0.7 for all models.

In summary, two observations stand out: First, omitting $10\,\%$ of training data (\texttt{900M-pre} vs.\ \texttt{1000M-pre}) does not significantly affect the performance on the formal linguistic competence captured by the BabyLM tasks. Second, adding only 1\,M additional words of interactive reinforcement learning after pretraining maintains those competences and even improves entity tracking.

\subsection{Storytelling}
\label{sec:results:storytelling}
%
How does interactive learning affect the learning dynamics of a small language model? We first explore the storytelling performance itself and the data efficiency of the learning setup. Next, we dive deeper into the learning dynamics of the individual storytelling criteria, the influence of the number of pretraining words and the interaction progress.  
\begin{figure}[htb]
    \centering
    \includegraphics[width=1\linewidth]{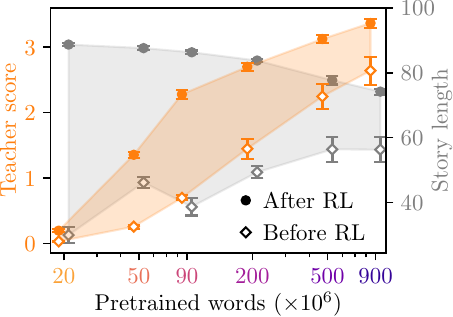}
    \caption{The effect of interactive reinforcement learning (RL) for models with increasing number of pretraining words on two variables: average teacher score (orange, left axis) and story length (gray, right axis). Error bars indicate the standard deviation of the first and last 20 batch averages, respectively. Orange and gray data points are slightly offset horizontally to avoid overlap.}
    \label{fig:PPO_start_finish}
\end{figure}

\paragraph{Storytelling skills} As shown in \Cref{fig:PPO_start_finish}, after the reinforcement learning (RL) interaction phase, the student models produce stories that are both longer and rated higher by the teacher. At first glance, this indicates that the models successfully learn to optimize the reward, which combines the teacher score and a bonus for story length. However, the extent of the improvement depends strongly on the amount of pretraining.

Specifically, models with more pretraining produce higher-scoring stories, both before and after RL: The 20\,M model initially produces short and after RL long stories that the teacher scores almost zero throughout. In contrast, the 90\,M and 200\,M models show the greatest increase in teacher score, while the most pretrained model, 900\,M, gains less from RL, although it ultimately achieves the highest absolute scores. Interestingly, the 900\,M model also produces the shortest stories after RL, despite earning the highest ratings, which suggests that it relies least on story length as a shortcut.

In \cref{app:sample_stories}, we provide a random sample of stories from the first, middle, and last third of interactions, as well as the best story, for the 90\,M and 900\,M models. The anthology of all stories produced by the models is available as a \href{https://huggingface.co/datasets/llm-slice/storytelling_anthology/tree/chck_900M}{Hugging Face dataset}.

\paragraph{Data efficiency} We find that interactive learning is remarkably data efficient: After RL, the 90\,M model receives an average teacher score of $2.3$ that outperforms that of the 500\,M model before storytelling interaction. Thus, 1\,M words of interactive learning achieve the same improvement as 410\,M extra words in pretraining. This result aligns with the findings of \citet{WuWZH2025} and \citet{zhao-etal-2023-babystories}, who demonstrate that LLMs learn with surprising efficiency in reinforcement learning. This robustness to sparse reward signals---such as the fixed student input in our setup---can be attributed to knowledge of the target domain acquired through sufficient pretraining. In our case, this finding agrees with our observation that a certain amount of pretraining is required before reinforcement learning can meaningfully enhance storytelling skills.

\begin{figure*}[htb]
    \centering
    \includegraphics[width=1\linewidth]{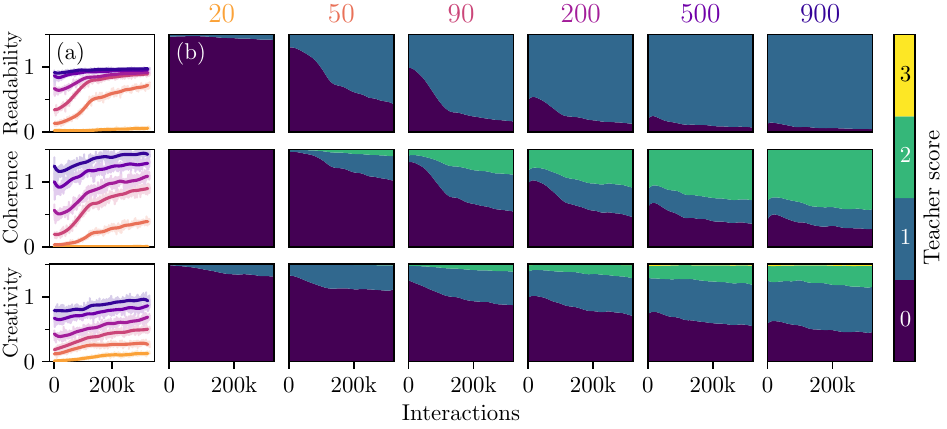}
    \caption{Teacher score over the course of RL interaction for models with increasing pretraining: 20\,M, 50\,M, \dots, 900\,M words. (a) Teacher scores by criterion. Shaded regions show averages per batch, solid lines are Gaussian-smoothed batch averages. (b) Distribution of teacher scores over time.}
    \label{fig:PPO_category}
\end{figure*}

\paragraph{Story quality}
We analyze the distribution of teacher scores across the criteria used for evaluating the student model's stories. \Cref{fig:PPO_category}~(a) shows the evolution of each criterion's score with the number of interactions for the six models with different amounts of pretraining. To emphasize underlying trends and filter out high-frequency fluctuations of the data, we apply a Gaussian filter.

Overall, the scores for all three criteria increase over time. As illustrated in \cref{fig:PPO_category}~(b), models with more pretraining perform better on all criteria. Notably, readability emerges as the hardest criterion, for which even the 900\,M model rarely attains two points, while performance in creativity and narrative coherence is substantially better across all models. The limited improvements on readability, which reflects superficial fluency, fit the observation from \cref{sec:results:blm_eval} that, for example, grammatical knowledge (as measured by \texttt{BLiMP}) is not much affected by the interactive RL, but creativity and coherence improve instead.

\cref{fig:PPO_category} also shows that the 20\,M model fails to achieve higher teacher scores except for a minor gain in creativity. Models pretrained for 90\,M and 200\,M words gain the most on all criteria, whereas more pretraining leads to diminishing returns in teacher scores.\footnote{Considering the entropy per word, see \cref{app:entropy}, we find that it is dominated by the amount of pretraining, with little change during interactive RL. This indicates that improvements in storytelling cannot simply be attributed to changes in output diversity.}

\begin{figure*}[htb]
    \centering
    \includegraphics[width=\linewidth]{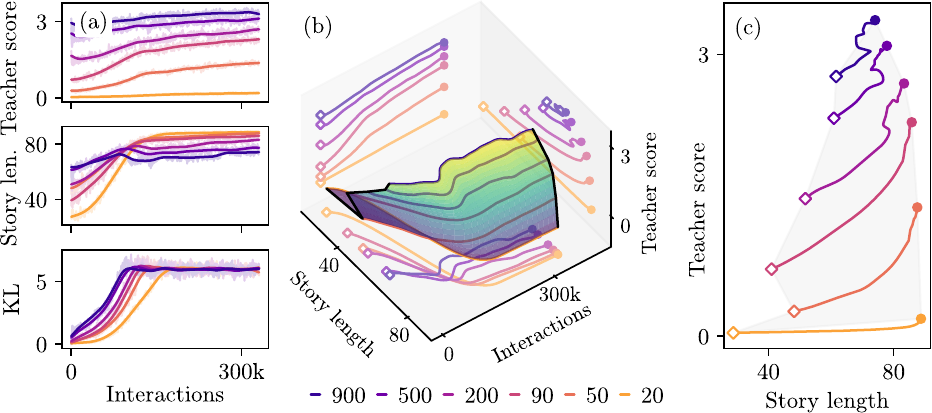}
    \caption{Learning dynamics of the reinforcement learning (RL). (a) Teacher score, story length, and KL divergence by interaction number. The shading shows the average per batch, the solid lines are Gaussian-smoothed batch averages. (b) Training trajectories visualized as a manifold with projections in the dimensions of story length (0 to 90 words) and teacher score (0 to 9 points) and number of interactions. (c) Trajectories in the phase space of teacher score and story length.}
    \label{fig:PPO_dynamics}
\end{figure*}
\paragraph{Learning dynamics}
\Cref{fig:PPO_dynamics}~(a) illustrates the evolution of teacher score, story length, and KL divergence over the number of interactions. Across all models, both teacher score and story length increase most rapidly until 100k interactions, after which improvements continue but at a slower pace. This deceleration is also reflected in \cref{fig:PPO_category}. KL divergence, which quantifies the similarity of the RL-trained model to its pretrained baseline, increases during early training and then stays constant around $\mathrm{KL} = 6$, a convergence determined by the adaptive KL scheduling of PPO. Deviating from this plateau would compromise the total reward signal, thus constraining policy updates. Notably, models with more pretraining, like 500\,M and 900\,M words, exhibit a decrease in story length after KL convergence before increasing again, potentially signaling a delayed adaptation of the model's reward prediction as these models adjust to changes in the slope of KL divergence.

\cref{fig:PPO_dynamics}~(b) combines the trajectories of the different models along three dimensions: story length, teacher score, and number of interactions. These trajectories define a surface, which we approximate with a one-dimensional linear interpolation (surface with blue to yellow gradient). The upper two diagrams in panel (a) correspond to projections of the trajectories, connecting the nonlinear effect of pretraining on the evolution of these variables.

\cref{fig:PPO_dynamics}~(c) completes the picture with a projection onto the plane of teacher score and story length, collapsing the dimension of interactions. This view reveals how models with different pretraining navigate the trade-off between story length and teacher score. The 20\,M model shows a limited slope, improving primarily in story length. This indicates a threshold: models pretrained on fewer than 50\,M words cannot leverage interactive feedback, which implies that some amount of pretraining is necessary for a viable reward signal. In contrast, the 90\,M and 200\,M models exhibit pronounced improvement in both dimensions. Models with even more pretraining like 500\,M and 900\,M display diminishing returns, consistent with \cref{fig:PPO_start_finish}. Overall, the 90\,M model benefits most from interactive learning.

\section{Conclusion}

Our experiments demonstrate that interactive feedback is highly data efficient for storytelling: With just 1\,M words of additional input, storytelling skills reach the equivalent of an additional 410\,M words of next-word prediction in pretraining. This result highlights the data inefficiency of next-word prediction and might explain why children acquire language with far less input than today's large language models.

We find that interactive reinforcement learning primarily enhances narrative coherence and creativity, while leaving surface-level fluency---measured by the BabyLM tasks---largely unchanged. An improvement in entity tracking aligns with the training objective focused on storytelling.

Our analysis reveals that models with less pretraining tend to exploit story length as a shortcut, whereas those with 90\,M and 200\,M words of pretraining benefit the most from interactive learning. Models with more pretraining suffer from diminishing returns from interaction. Notably, we identify a threshold: between 20\,M and 50\,M words of pretraining are necessary for the model to benefit from interactive reinforcement learning. Examining the nature of this threshold and its parallels to language acquisition in children presents an intriguing avenue for future research.

\section*{Limitations}
While storytelling RL is highly data efficient, it is by no means computationally efficient: RL on 1\,M input words took 20 GPU hours per model, because it involves generating 20\,M words of student output for the stories. For comparison, 900\,M words of pretraining amounted to less than 10 GPU hours.

Moreover, our analysis focuses on the learning dynamics. We leave a detailed study of the student stories---how content, register, vocabulary, and syntax evolve through interaction---for future work. Mechanistic interpretability methods could also provide insights into how training affects internal model representations.

Furthermore, we weight the three evaluation criteria of the teacher equally, but these weights can be adapted during RL to implement a form of curriculum learning.

Our teacher rewards serve as a heuristic for story quality. Further validation using benchmarks like \texttt{OpenMEVA} \cite{GuanZFLD2021} or human annotations would strengthen this approach.

We used a fixed input for story generation, but more diverse corpora (e.g., BabyLM \cite{CharpentierCCGH2025}, TinyStories \cite{EldanL2023}, or WritingPrompts \cite{FanLD2018}) could affect learning outcomes; each with its own tradeoffs regarding narrative content and diversity.

\section*{Ethics statement}

Importantly, computational language models are not faithful representations of human cognition and should not be anthropomorphized. Rather, they are tools for informing hypotheses about language learning, which should ultimately be tested on human studies.

While the BabyLM challenge targets more sustainable training regimes, model development still requires considerable computing resources. Model development and final training took about 140\,kcore-hours in total. Pretraining took 2 hours on 4 A100 GPUs. RL learning took 20 hours on 1 A100 GPU for each of the six RL models (5 - 10 kcore-hours per model).

\section*{Acknowledgements}

We thank the reviewers for their input.
We thank Eva Beck for helpful discussions.
Lisa Beinborn’s research is partially supported by an \emph{Impulsprofessur} grant from the \emph{zukunft.niedersachsen} program and by a VENI grant (Vl.Veni.211C.039) from the Dutch National Science Organisation (NWO).
The authors gratefully acknowledge computing time provided to them at the GWDG HPC cluster.
\bibliography{custom}

\appendix
\clearpage
\section{Evaluation}
\label{app:BLM_evals}

\begin{table*}[htb]
    \centering
    \small
    \begin{tabular}{lllll}
        \toprule
       Setting & Dataset & Prediction Task & Evaluation Metric & Reference \\
        \midrule
        \multirow{9}{*}{\rotatebox{90}{Zero-shot}} & \texttt{BLiMP} &Grammatical acceptability & Accuracy & \citet{WarstadtPLMP2020}\\
        & \texttt{Suppl.}   & Discourse acceptability & Accuracy & \citet{WarstadtMCWZ2023} \\
        &  \texttt{EWOK}     & Conceptual knowledge & Accuracy  &   \citet{IvanovaSLKR2025} \\
        &  \texttt{COMPS}     & Property knowledge & Accuracy  &   \citet{misra-etal-2023-comps} \\
        &    \texttt{WUG-A}      & Morphol.\ generalization (adj.) & Spearman's $\rho$ &    \citet{weissweiler-etal-2023-counting}\\
        &    \texttt{WUG-P}      & Morphol.\ generalization (verbs) & Spearman's $\rho$ &   \citet{HofmannWMSP2024}\\
        &    \texttt{AoA}       & Age of acquisition &Pearson's $\rho$  &   \citet{chang-bergen-2022-word} \\
        &    \texttt{Eye-T}    & Eye-tracking fixations & Squ.\ partial corr.\ $\mathrm{p}R^2$ &  \citet{DeVardaMA2023} \\ 
        &    \texttt{SPR}      & Reading times & Squ.\ partial corr.\ $\mathrm{p}R^2$ &    \citet{DeVardaMA2023}\\
        &    \texttt{ET}       & Entity Tracking & Accuracy &   \citet{KimS2023} \\
        \midrule
        \multirow{ 7}{*}{\rotatebox{90}{Fine-tuning}} & \texttt{BoolQ}    & Question answering  & Accuracy &\citet{clark-etal-2019-boolq} \\
        &    \texttt{MultiRC}  & Question answering & Accuracy &  \citet{khashabi-etal-2018-looking}  \\
        &       \texttt{MNLI}  & Natural language inference & Accuracy &  \citet{williams-etal-2018-broad}\\ 
        &    \texttt{RTE}      & Entailment & Accuracy &  \citet{bentivogli2009fifth}  \\
        &    \texttt{MRPC}    & Paragraph identification & Accuracy &   \citet{dolan-brockett-2005-automatically} \\
        &    \texttt{QQP}      & Question similarity  & Accuracy &   \href{https://quoradata.quora.com/First-Quora-Dataset-Release-Question-Pairs}{Iyer et al., (2017)} \\
        &    \texttt{WSC}      & Coreference resolution &Accuracy  &   \citet{levesque2012} \\
        \bottomrule
    \end{tabular}
    \caption{Overview of evaluation datasets in the BabyLM pipeline.}
    \label{tab:evaluation_setup}
 \end{table*}
 
We use the \href{https://github.com/babylm/evaluation-pipeline-2025}{evaluation pipeline} of the 2025 BabyLM Challenge \cite{CharpentierCCGH2025}. In \cref{tab:evaluation_setup}, we provide an overview of the evaluation data.

\section{Entropy} \label{app:entropy}

\begin{figure}[htb]
    \centering
    \includegraphics[width=1\linewidth]{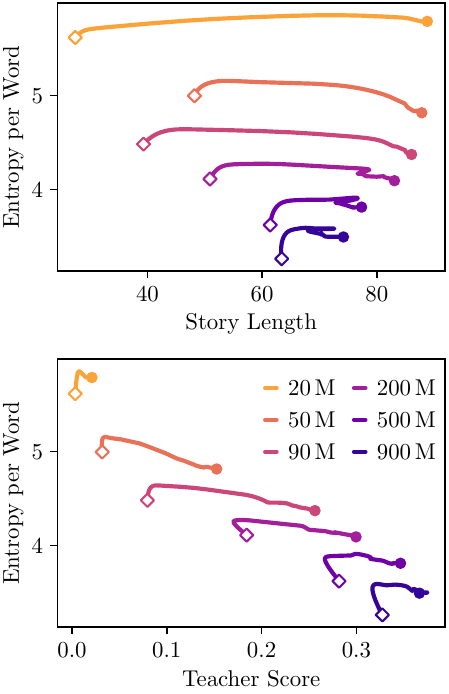}
    \caption{Entropy per word by story length and teacher score during interactive RL for different amounts of pretrained words. An empty diamond marks the start of a trajectory, a filled circle the end.}
    \label{fig:entropy}
\end{figure}

Figure \cref{fig:entropy} shows that the average entropy per word increases slightly at the beginning of training, staying mostly constant until the end of reinforcement learning, but the entropy is otherwise not substantially correlated with story length or teacher score. Pretraining, on the other hand, has a strong influence on the entropy per word.

\vspace{1cm}
\section{Sample stories} \label{app:sample_stories}

Best story by reward and example stories---randomly sampled from the first, second, and last third of RL training---are listed in \cref{tab:example-stories-90m} for 90\,M pretrained words and \cref{tab:example-stories-900m} for 900\,M pretrained words. Interestingly, the best story for 900\,M pretrained words is a meta-story---a story about a story---that directly appeals to the teacher evaluation by describing a \enquote{great story}. Each model produced about 20\,M words during RL training, which amounts to about 50,000 pages. The full anthology is available as a \href{https://huggingface.co/datasets/llm-slice/storytelling_anthology/tree/chck_900M}{Hugging Face dataset}.
\newpage
\begin{table}[htbp]
    \centering
    \setlength{\tabcolsep}{3pt}
    \begin{tabular}{p{1\linewidth}}
    \toprule
    \small (1, 2, 3) \quad \textbf{Length:} 82 \quad \textbf{Reward:} 0.71 \quad \textbf{\#} 318,619 \\
    \small when it was given to the sun.  Did he tell you that I wrote a letter, then that the moon was not frightened of telling her story? She introduced her red coat into the eyes of his safety. Perhaps she loved that, though she had seen sad old laughter. She now told her primary harmon they in spite of the name as she had never seen a star light thunder escort. She found herself there a dragon doing her best at it \\
    \midrule
    \small (0, 0, 0) \quad \textbf{Length:} 25 \quad \textbf{Reward:} 0.07 \quad \textbf{\#} 43,434 \\
    \small she will possess me a descend-weller repwined feast. Of this of the Saxon divinity and for fifty months will occur; '' Mr. March reppainted .</s> \\
    \midrule
    \small (1, 1, 1) \quad \textbf{Length:} 67 \quad \textbf{Reward:} 0.43 \quad \textbf{\#} 217,857 \\
    \small hitherto hathine before twenty-three of people had lost their Senator Ragnzan leave in the decision, formally on the fact that he would make-up payment.  After Commonwealth negotiated against a white elector offenders, Captain Spndez Junpped with the alliance and denied the suggestion which was only a vote from him: he seems to men such as England and Wales, but he said not to have been a party, \\
    \midrule
    \small (1, 2, 1) \quad \textbf{Length:} 71 \quad \textbf{Reward:} 0.52 \quad \textbf{\#} 321,361 \\
    \small which whenever it may not be trusted.  'Cause she really would feel about that she might have recognized her lyrics.  I encourage her.  The love is not yet permitted to pronouncemy to the contrary."  She was in a state that it was a small, sweet gift.  She kept pieces of the discussion, and she heard her dear face.  Her newence was based upon her key to the audit that she had \\
    \bottomrule
    \end{tabular}
    \caption{Examples of generated stories from the 90\,M model: overall best story (top) ranked by teacher and length reward, plus random samples (seed 42) from the first, second, and last third. Prompt: \enquote{\emph{Let me tell you a long, magical tale. Once upon a time, in a faraway land}}.}
    \label{tab:example-stories-90m}
\end{table}

\begin{table}[htbp]
    \centering
    \setlength{\tabcolsep}{3pt}
    \begin{tabular}{p{1\linewidth}}
    \toprule
    \small (2, 3, 3) \quad \textbf{Length:} 76 \quad \textbf{Reward:} 0.85 \quad \textbf{\#} 8,821 \\
    \small everything seemed as if it were a fairy tale—so much as you have told us a great story that we cannot bear to believe. There was something, however, very different from the description of the whole story, most surprising. If the story reached out, then, and it was always the story at the end, it was always something inexpressibly a story that made a great deal of sense. But it was all a mistake and wonder \\
    \midrule
    \small (1, 2, 2) \quad \textbf{Length:} 76 \quad \textbf{Reward:} 0.61 \quad \textbf{\#} 43,434 \\
    \small there lived old men like me, the flowers of Jed serene rose from every mountain and valley, and they grew wildly from one to another. The inhabitants of those under one tree and the branches in the other, had been bewitched in their own world by their native dwelling; but having enabled them to approach them they had for good to take an active place among mankind. They resided many weary people, but had always a \\
    \midrule
    \small (1, 0, 0) \quad \textbf{Length:} 63 \quad \textbf{Reward:} 0.26 \quad \textbf{\#} 217,857 \\
    \small there lived an unquenchable king. But cannot the words wear off, and for ten days he was forced to stake his head on one native bird-cage, while the hunters came in for us the fish-skin preservation. He was a dread of poor little war-birds, and a more likeable wickedness so lonesome in proportion to his cruel fangs as a young bird devouring li \\
    \midrule
    \small (1, 0, 0) \quad \textbf{Length:} 67 \quad \textbf{Reward:} 0.27 \quad \textbf{\#} 321,361 \\
    \small immature and upland boy, he was dazzled by the tremendous overlooks of his race had lighted. He met a spirit who had been there all the day to bespeak in the midst of many years, and answered: "Hear him, Don Carlos, from there he lent it to reality; He is a different kind of drunken-looking man; I consider him much creamered after his teeth." In the same \\
    \bottomrule
    \end{tabular}
    \caption{Examples of generated stories from the 900\,M model: overall best story (top) ranked by teacher and length reward, plus random samples (seed 42) from the first, second, and last third.}
    \label{tab:example-stories-900m}
\end{table}

\clearpage
\onecolumn
\section{Evaluation instructions} 
\label{app:teacher_prompt}

\begin{minipage}{\textwidth}
\begin{tcolorbox}[colback=gray!10, colframe=black!80]
\small
\begin{verbatim}
You are a helpful teacher grading a student story. Be nice!
Only evaluate the student story itself, not the story prompt.
Given the student's word limit of about 80 words,
score the story on each of these three categories separately
on a scale from 0 to 3,
where 0 is the worst and 3 is the best.

Readability:
0 - Frequent and severe grammar errors; difficult to understand.
1 - Noticeable grammar errors; mostly understandable.
2 - Few minor grammar errors; well-formed overall.
3 - Correct grammar; well written.

Narrative Coherence:
0 - No story: completely incoherent or too short.
1 - No logical flow, confusing narrative.
2 - Mostly coherent story and not cut off.
3 - Coherent and logically structured story.

Creativity:
0 - Dull or incomprehensible.
1 - Somewhat creative; mostly predictable.
2 - Fairly creative and engaging.
3 - Highly original, imaginative, and engaging.

If the student story is empty ("") or less than a full sentence,
you must give the score 0 0 0!

Provide your scores, separated by single spaces, in the format:
Readability, Narrative, Creativity = _ _ _

Respond ONLY with this sequence of three numbers
without any extra text or explanation.

Story Prompt:
`{{story_prompt}}`

Student Story:
"`{{student_completion}}`"

Readability, Narrative, Creativity =
\end{verbatim}
\end{tcolorbox}

\vspace{2cm}

\noindent
\begin{minipage}{\textwidth}
    \centering
    \small
    \begin{tabular}{llllll}
        \toprule
         \textbf{Source} & \textbf{Ratio} & \textbf{Domain} & \textbf{Reference} \\
         \midrule 
         \href{http://www.natcorp.ox.ac.uk/}{BNC} & 8\% & Dialogue & \citet{BNCConsortium2007} \\
         CHILDES & 29\% & Dialogue, child-directed & \citet{MacWhinney2014} \\
         \href{https://github.com/pgcorpus/gutenberg}{Proj.\ Gutenberg} & 26\% & Fiction, nonfiction & \citet{GerlachF2018} \\
         \href{https://opus.nlpl.eu/OpenSubtitles-v2018.php}{OpenSubtitles} & 20\% & Dialogue, scripted & \citet{LisonT2016} \\
         \href{https://www.loc.gov/item/2019205402/}{Simple Eng.\ Wiki.\ } & 15\% & Nonfiction & -- & \\
         \href{http://compprag.christopherpotts.net/swda.html}{Switchboard} & 1\% & Dialogue & \citet{GodfreyHM1992}, \citet{StolckeRCSB2000} \\
        \bottomrule
    \end{tabular}
    \captionof{table}{Composition of the \href{https://osf.io/ryjfm/}{BabyLM corpus}.}
    \label{tab:BLM_corpus}
\end{minipage}
\end{minipage}

\clearpage

\twocolumn

\section{Model parameters}\label{app:model_parameters}

\paragraph{BabyLM corpus}

The composition of the BabyLM corpus is listed in \cref{tab:BLM_corpus}. It comprises 100\,M words, of which we use 90\% for pretraining and tokenization.

\subsection{Pretraining}\label{app:model_parameters:pretraining}

\paragraph{Model and training} The model parameters are listed in \cref{tab:pretrain_parameters}. The vocab size of the tokenizer is 16,000 to match the baseline $\texttt{1000M-pre}$ and the interaction baseline \texttt{SimPO}, which have vocab size 16,384. We use different values for seed, batch size, gradient accumulation, and learning rate compared with the baselines.

\begin{table}[htb]
    \centering
    \small
    \begin{tabular}{lr}
        \toprule
        \textbf{Hyperparameter} & \textbf{Value} \\
        \midrule
        Number of epochs & 10 \\
        Context length & 512 \\
        Batch size & 16 \\
        Gradient accum.\ steps & 4 \\
        Learning rate & 0.0005 \\
        Number of steps & 211,650 \\
        Warmup steps & 2,116 \\
        Gradient clipping & 1 \\
        Seed & 42 \\
        Optimizer & AdamW \\
        Optimizer $\beta_1$ & 0.9 \\
        Optimizer $\beta_2$ & 0.999 \\
        Optimizer $\varepsilon$ & $10^{-8}$ \\
        Tokenizer & ByteLevelBPE \\
        Tokenizer vocab size & 16,000 \\
        Tokenizer min.\ frequency & 2 \\
        \bottomrule
    \end{tabular}
    \caption{Hyperparameters used for pretraining.}
    \label{tab:pretrain_parameters}
\end{table}

\subsection{Reinforcement learning}\label{app:model_parameters:RL}

See \cref{tab:hyperparameters-ppo}.

\begin{table}[htb]
    \small
    \centering
    \begin{tabular}{l r}
        \toprule
        \textbf{Parameter}              & \textbf{Value} \\
        \midrule
        Student context length          & 512   \\
        Seed                            & 42    \\
        Batch size                      & 360   \\
        Student sampling temp.\         & 1     \\
        Top $k$                         & 0     \\
        Top $p$                         & 1     \\
        Max.\ new tokens (student)      & 90    \\
        Teacher model                   & Llama~3.1 \\ 
                                        & 8B~Instr.\ \\
        Teacher context length          & 1,024  \\
        Student sampling temp.\         & 0.2   \\
        Max.\ new tokens (teacher)      & 6     \\
        Gradient acc.\ steps            & 1     \\
        Adapt.\ KL control              & True  \\
        Init.\ KL coef.\                & 0.2   \\
        Learning rate                  & $1\times 10^{-6}$\\  
        Student input limit             & 1\,M~words \\
        \bottomrule
    \end{tabular}
    \caption{PPO Training Hyperparameters. Other parameters defaults of TRL 0.9.4.}
    \label{tab:hyperparameters-ppo}
\end{table}

\subsection{Fine-tuning}\label{app:model_parameters:fine-tuning}

See \cref{tab:hyperparameters-fine-tuning}.

\begin{table}[htb]
    \centering
    \small
    \begin{tabular}{lr}
        \toprule
        \textbf{Hyperparameter} & \textbf{Value} \\
        \midrule
        Number of Epochs & 10 \\
        Batch Size & 16 \\
        Learning Rate & $3 \times 10^{-5}$ \\
        Warmup percentage & 6\,\% \\ 	
        Optimizer & AdamW \\
        Weight decay & 0.01 \\
        Scheduler & cosine \\
        Dropout & 0.1 \\
        \bottomrule
    \end{tabular}
    \caption{Hyperparameters used for fine-tuning.}
    \label{tab:hyperparameters-fine-tuning}
\end{table}

\section{BabyLM evaluation results}

See \cref{tab:BLM_eval_results}.

\begin{table}[htb]
    \centering
    \small
    \setlength{\tabcolsep}{2pt}
    
    \begin{tabular}{l
                    S[table-format=2.2]
                    S[table-format=2.2]
                    S[table-format=2.2]
                    S[table-format=2.2]}
    \toprule
    \textbf{Task} & \textbf{1000M-pre} & \textbf{SimPO} & \textbf{900M-pre} & \textbf{900M-RL} \\
    \midrule
    \texttt{BLiMP}   & 74.88 & 72.16 & 77.52 & {\bfseries 77.53} \\
    \texttt{Suppl.}  & {\bfseries 63.32} & 61.22 & 56.62 & 56.72 \\
    \texttt{EWOK}    & 51.67 & {\bfseries 51.92} & 51.36 & 51.41 \\
    \texttt{COMPS}   & {\bfseries 56.17} & 55.05 & 55.20 & 55.18 \\
    \texttt{ET}      & 31.51 & 28.06 & 30.34 & {\bfseries 33.11} \\
    \texttt{GLUE}    & 52.18 & 50.35 & {\bfseries 53.14} & 52.46 \\
    \bottomrule
    \end{tabular}
    
    \vspace{0.4em}
    
    \begin{tabular}{l
                    S[table-format=1.3]
                    S[table-format=1.3]
                    S[table-format=1.3]
                    S[table-format=1.3]}
    \toprule
    \textbf{Task} & \textbf{1000M-pre} & \textbf{SimPO} & \textbf{900M-pre} & \textbf{900M-RL} \\
    \midrule
    \texttt{WUG-A}   & 0.502 & 0.510 & 0.701 & {\bfseries 0.711} \\
    \texttt{WUG-P}   & 0.073 & {\bfseries 0.179} & 0.042 & 0.045 \\
    \texttt{AoA}     & 0.053 & 0.074 & {\bfseries 0.080} & {\bfseries 0.080} \\
    \texttt{Eye-T}   & 0.079 & {\bfseries 0.091} & 0.003 & 0.002 \\
    \texttt{SPR}     & 0.032 & {\bfseries 0.035} & 0.000 & 0.000 \\
    \bottomrule
    \end{tabular}
    
    \caption{BabyLM task scores for the four models from \cref{fig:blm_eval}. 
    Accuracy metrics are reported as percentages, WUG-A/P as Spearman’s~$\rho$, AoA as Pearson’s~$\rho$, Eye-T and SPR as partial correlations~$\mathrm{p}R^2$. Bold indicates the best model for each task.}
    \label{tab:BLM_eval_results}
\end{table}

\end{document}